\documentclass[10pt,twocolumn,letterpaper]{article}

\usepackage{cvpr}
\usepackage{times}
\usepackage{epsfig}
\usepackage{graphicx}
\usepackage{amsmath}
\usepackage{amssymb}
\usepackage{color,xcolor}

\definecolor{Tianlong_color}{rgb}{0.858, 0.188, 0.478}

\usepackage[pagebackref=true,breaklinks=true,letterpaper=true,colorlinks,bookmarks=false]{hyperref}

\cvprfinalcopy 

\def\cvprPaperID{0001} 

\ifcvprfinal\fi
\begin{document}

\title{Focus Longer to See Better: \\ Recursively Refined Attention for Fine-Grained Image Classification}

\author{Prateek Shroff\textsuperscript{1}, Tianlong Chen\textsuperscript{1}, Yunchao Wei\textsuperscript{2}, Zhangyang Wang\textsuperscript{1}\\
\textsuperscript{1}Texas A\&M University,\textsuperscript{2}University of Technology Sydney\\
{\tt\small \{prateek.shroff,wiwjp619,atlaswang\}@tamu.edu,wychao1987@gmail.com}
}

\maketitle
\thispagestyle{empty}

\begin{abstract}
Deep Neural Network has shown great strides in the coarse-grained image classification task. It was in part due to its strong ability to extract discriminative feature representations from the images. However, the marginal visual difference between different classes in fine-grained images makes this very task harder. In this paper, we tried to focus on these marginal differences to extract more representative features. Similar to human vision, our network repetitively focuses on parts of images to spot small discriminative parts among the classes. Moreover, we show through interpretability techniques how our network focus changes from coarse to fine details. Through our experiments, we also show that a simple attention model can aggregate (weighted) these finer details to focus on the most dominant discriminative part of the image. Our network uses only image-level labels and does not need bounding box/part annotation information. Further, the simplicity of our network makes it an easy plug-n-play module. Apart from providing interpretability, our network boosts the performance (up to 2\%) when compared to its baseline counterparts. Our codebase is available at \url{https://github.com/TAMU-VITA/Focus-Longer-to-See-Better}
\end{abstract}

\section{Introduction}
\begin{figure}[t]
\includegraphics[width=\linewidth]{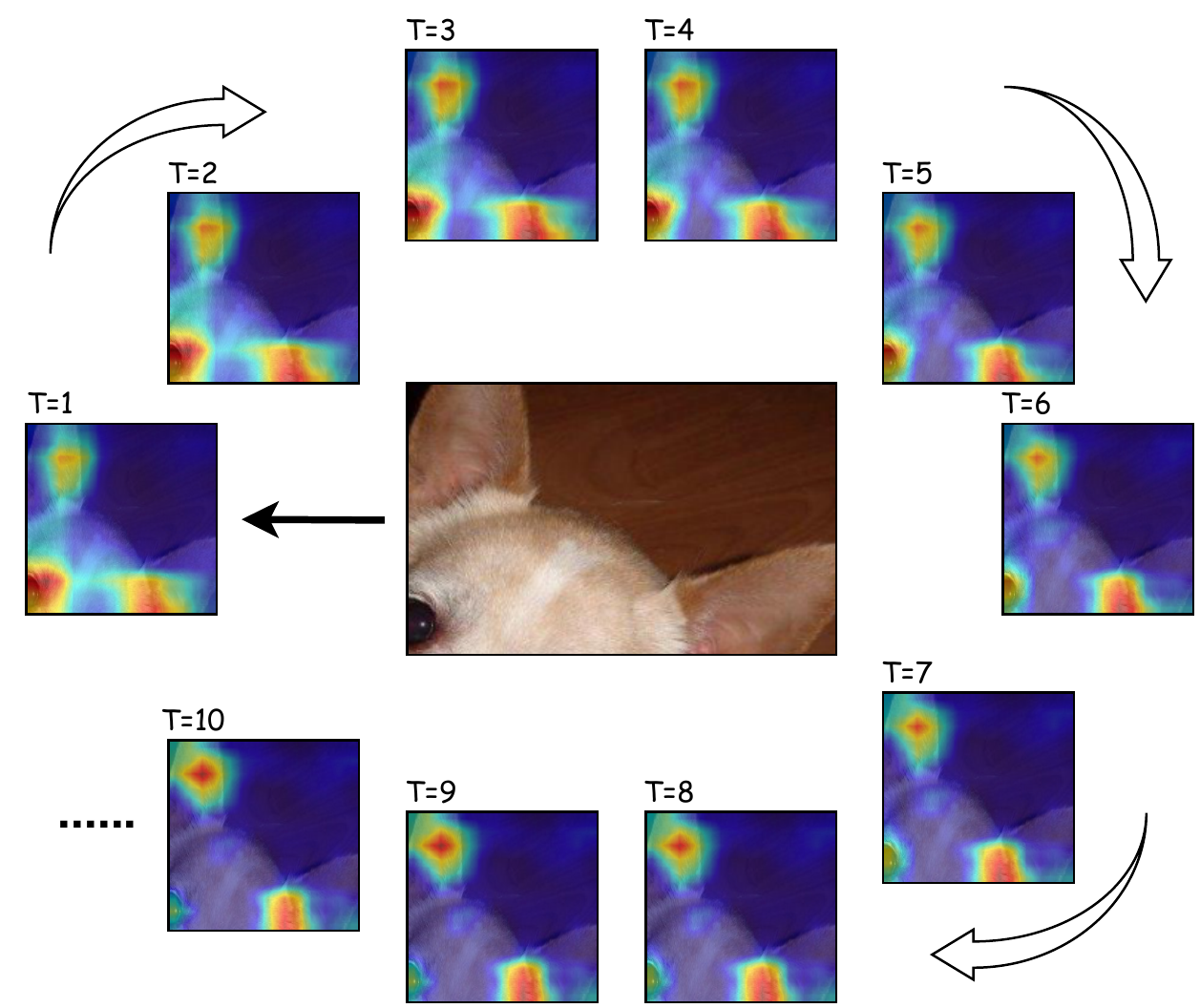}
\caption{Example of a center patch of an image. The heat-maps around the image visualize the changes in attention, as we look longer at an image from temporal step ($\mathrm{T}$) from $\mathrm{1}$ to  $\mathrm{10}$. Looking recurrently at an image helps our model to progressively spot finer details like pointy ears from the coarser head region.}
\label{fig:tisser}
\end{figure}

Fine-grained image classification \cite{1} has been an active research area that recognizes sub-categories within some meta-category \cite{cub,dogs,car}. This problem differs from generic image classification due to very small inter-class and large intra-class variations among the classes. This makes the task very challenging, as the recognition should be able to localize these fine variations and then represents these marginal visual differences. Though deep neural networks \cite{1} have shown astounding performance in generic image classification \cite{ILSVRC15}, reaching similar-level performance on fine-grained recognition remains a challenge. 

Deep learning approaches for fine-grained classification fall into two separate paradigms: localization-classification network \cite{look, macnn, ge2019weakly}, and end-to-end feature encoding \cite{bilinear, kernel, hfv}. The first category makes use of a separate localization network along with the classification network. The localization network is used to localize the discriminative image regions/parts. In order to localize these fine changes, earlier work \cite{mgcnn, FCNN} has relied on human-annotated bounding box/par annotations (eg head, wings, feather color). But, all these human-based manual annotations make the process quite intensive, laborious, and subjective. Also, the manual annotation may be possible for small scale datasets like CUB200-2011 \cite{cub} , Stanford dataset \cite{dogs, car} but not feasible for large-scale image dataset say ImageNet \cite{ILSVRC15}. Convolution neural networks (CNNs) were hence leveraged for weakly supervised part-learning with category-labels only, assuming no dependencies on bounding box/part annotations \cite{2,3,4}.

In the localization-classification network, the localization subnetwork focuses on learning the objects parts shared among the same classes while the classification subnetwork extracts discriminative features from these localize objects to make them different among classes. This complementary network architecture requires separate losses \cite {mgcnn, look} and tends to be computationally expensive.

The second category is to encode higher-order statistics of convolutional feature maps to enhance the feature presentation of the image \cite{bilinear, compact, low, u}. One of the first works in this category was the use of Bilinear CNNs \cite{bilinear} which computes pairwise feature interactions by two independent CNNs to capture the local differences in the image. Another work \cite{fisher} proposed to encode CNN representation with Fisher Vector representation giving much superior performance on several datasets. But using higher-order dynamics makes the network less human-interpretable when compared to the localization-classification sub-network. 

To overcome the above-mentioned challenges, we propose a novel attention-based recurrent convolutional neural network for fine-grained image classification. Our network recursively attends from coarse to the finer region of image or parts of the image to focus on the discriminative region more finely.  Our model is simple, computationally inexpensive, and interpretable. Our motivation is that by processing an image or a part of the image recursively, we can focus on most discriminative details by continuously removing insignificant ones and other background noises. Further,  by aggregating the finer regions from the image via suitable attention we can pinpoint the most discriminative region in the image. Additionally, the module is plug-and-play which greatly enhances its scalability and usability. 

Our network consists of a weakly supervised patch extraction network which extracts different patches corresponding to an image. Another network attends to each patch by recurrently processing it via LSTMs. We use uni-directional stacked LSTMs to recurrently pass the patch through the time steps of LSTMs.  Then, an attention layer is used to aggregate the finer representation from the output of the LSTMs. We append this network to the baseline image classifier giving way to a two-stream architecture. To leverage the power of ensembles, the representative features are fused and then passed to the end classifier. 

Our contributions can be summarized as the following:
\begin{itemize}
    \item {We propose a novel recurrent attention network which progressively attends to and aggregate finer image details, for more discriminative representations.}
    \item{ We show through various ablation studies the human interpretability of our attentions and features.}
    \item{ We conduct experiments on two challenging benchmarks (CUB200-2011 birds \cite{cub}, Stanford Dogs \cite{dogs}), and show performance boosts over the baselines.} 
\end{itemize}


\section{Related Work}



\textbf{Fine-grained Feature Learning.} Learning discriminative features have been studied extensively in the field of image recognition and also for fine-grained classification. Due to the great success of deep learning, powerful deep convolutional based features \cite{1,resnet,densenet,googlenet} forms the backbone for most of the recognition tasks. This has shown a great boost in performance when compared to hand-crafted features. To model subtle difference, a bilinear structure \cite{bilinear} is used to compute pairwise differences. The use of boosting to combine the representation of multiple learners also helps to improve classification accuracy \cite{moghimi2016boosted}. Additionally, second-order information also helps in fine-grained feature extraction. Pooling methods that utilize second-order information \cite{s,u} have proven to enhance the extraction of more meaningful information.

\textbf{Interpretable Deep Models for Fine-grained Recognition.} Given the subtle differences between fine-grained categories, it becomes imperative to focus on and extract meaningful features from them. There has been extensive research \cite{zhou2016learning, pscnn,atten2,neural,thisLook} to develop interpretable models that visualize regions attended by the network. In \cite{zhou2016learning} , Class Activation Maps (CAMs) are used to provide object-level attention but not providing much finer discriminative details. Over time, there have been variants developed \cite{gradcam, inter1}, that explores the backward propagation to identify salient image features. In \cite{pscnn,atten2,neural}, attention is at a finer level and focus more on the parts of the object rather than the whole body/object. In \cite{thisLook}, the authors associate the prototypical aspect with the object part to reason out the classification prediction for an image. Our network uses a simple approach based on \cite{gradcam} to visualize the fine attention areas in the patches. 

\textbf{Attention.} Attention has been incorporated in visual related tasks from a long time \cite{show,recur,active,top,opt}. Attention models are aimed at identifying discriminative image parts that are most responsible for recognition. We follow on the same methodology of the visual-attention model to aggregate the output of LSTMs to have weighted attention to the most discriminative patch/part of the image. In \cite{ge2019weakly}, the author uses weakly supervised model to generate different patches of the same image containing different parts of images. We used a similar approach to extract patches from the images which is further used to look for finer details. This method does not use any external information like part annotations/bounding box information.

\begin{figure*}
\includegraphics[width=\linewidth]{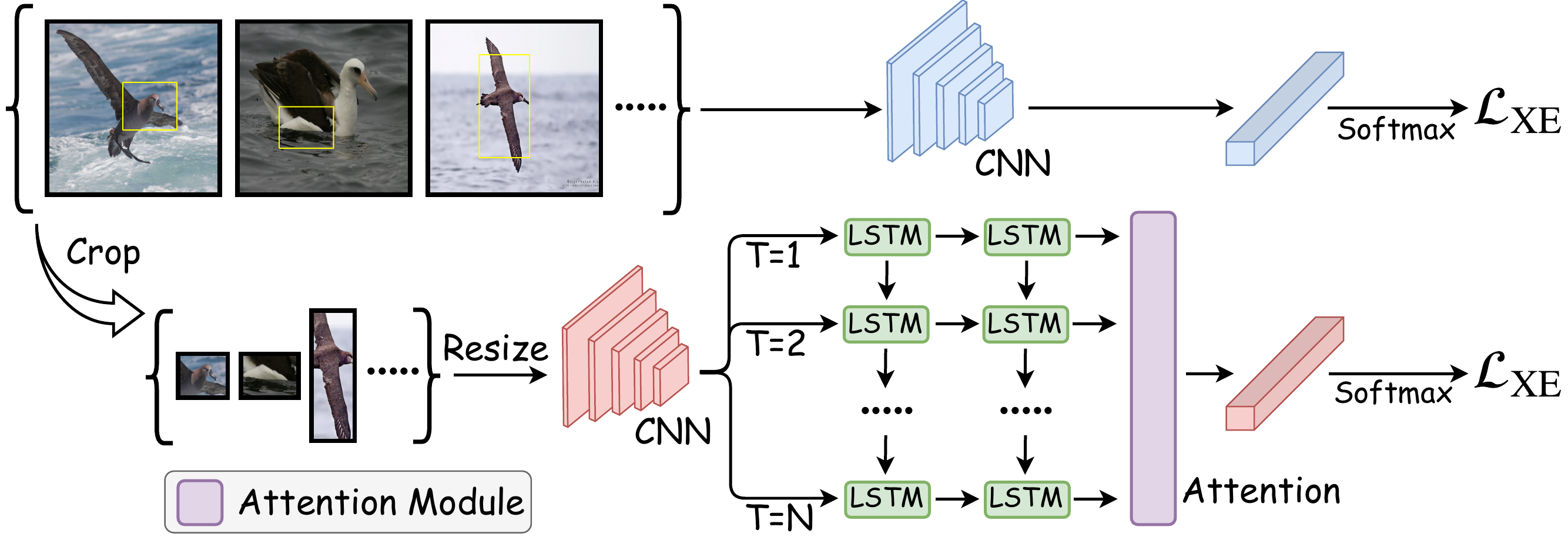}
\caption{The pipeline of our two-stream architecture. The global stream (on the top) processes the entire image to provide global representation. While the local stream (on the bottom) processes a certain region of the image (say, patch). The features are generated by recurrently passing the patch through stacked LSTMs followed by an attention layer. Finally, the whole architecture is optimized via cross-entropy loss at each output. }
\label{fig:arch}
\end{figure*}
\section{Our Proposal}
Given an image $\mathrm{I}$ and its corresponding label $\mathrm{c}$, our network aims to look longer via recurrently iterating through a patch of an image to extract more fine-grained information. A bottom-up weakly supervised object detection approach is used to extract meaningful patches (parts of the images). This network uses only the category level labels and does not use any part annotations or bounding box information. Further, a two-stream feature extractor is used to extract global and object-level feature representations to boost the classification accuracy.


\subsection{Two-Stream Architecture}
Once we get a set of patches for each training image $\mathcal{I}$, we randomly select a patch  $\mathcal{P}_\mathrm{i}$ from the set of patches $\mathcal{P}$ obtained. Hence, the input to the two-stream architecture consists of image $\mathcal{I}$ and a patch  $\mathcal{P}_\mathrm{i}$ defined by a pair of coordinates 
\begin{equation}
    [(\mathrm{x_{i}^{tl}},\mathrm{y_{i}^{tl}}), (\mathrm{x_{i}^{br}}, \mathrm{y_{i}^{br}})]
\end{equation}
where  $\mathrm{tl}$ and $\mathrm{br}$ represent top-left and bottom-right. The pair of coordinates denote top-left and bottom-right corners of the box over the part of an image. Assuming top-left corner in the original image as the origin of a pixel coordinate system, x-axis and y-axis is defined from left-to-right and top-to-bottom respectively.

 As shown in figure \ref{fig:arch}, there are two streams in the architecture. The top stream consists of a convolution-based feature extractor followed by the classification layer. The second stream takes patch from images and extracts feature presentations via CNN. These features are recurrently passed through LSTMs to get better and finer representations focusing on fine discriminative regions within the patch. These finer patches are weight-aggregated to form a single most discriminative representation. Specific details about the architecture are shared in the following sections. 

\paragraph{Global Stream}
Given an input image $\mathcal{I}$, we first extract deep features by passing the image through a convolution neural network. The neural network is pretrained on ImageNet \cite{ILSVRC15}. The extracted representations can be written as $\textbf{W}_\mathrm{g}$ * $\mathcal{I}$, where $\textbf{W}_\mathrm{g}$ denotes the representative weight of the whole neural network and * denotes all the convolution, pooling, and non-linear functions performed on the input image. The features are further passed through a softmax layer which outputs a probability distribution over fine-grained categories. Mathematically, 
\begin{equation}
    \mathcal{\textbf{G}}_\mathrm{I} =  \mathcal{F}( \textbf{W}_\mathrm{g} * \mathcal{I} )
\end{equation}

where $\textbf{G}_\mathrm{I}$ represents global representation for image and  $\mathcal{F}$(.) denotes the Global Pooling Layer (GAP) \cite{nin} followed by a fully connected softmax layer which transforms the deep features into probabilities. The global stream is used to extract global representative features of the images.  The reasons for including this simple branch are two-fold. First, to provide global information to the network during the training since the patches/parts of the object extracted focus on the object itself. Second, it provides a simple baseline over which our local stream can be added, demonstrating the plug-n-play functionality of our main contribution.

\paragraph{Local Stream}
The output of weakly supervised patch extraction framework is dominant parts of an image as $\mathcal{P}$ = [$\mathcal{P}_\mathrm{1}$, $\mathcal{P}_\mathrm{2}$, $\mathcal{P}_\mathrm{3}$, ..., $\mathcal{P}_\mathrm{n}$], where each $\mathcal{P}_\mathrm{i}$ could be defined as a pair of coordinates of the bounding box for a region of an image. The regions are cropped from the entire image as shown in the figure \ref{fig:arch}. The set of cropped image regions can be denoted as  $\mathcal{I}( \mathcal{P})$ = [ $\mathcal{I}(\mathcal{P}_\mathrm{1})$, $\mathcal{I}(\mathcal{P}_\mathrm{2})$, $\mathcal{I}( \mathcal{P}_\mathrm{3})$, ..., $\mathcal{I}(\mathcal{P}_\mathrm{n})$]. Once a region $\mathcal{I}(\mathcal{P}_\mathrm{i})$ (say $\mathrm{i}_{th}$ patch) is cropped from image  $\mathcal{I}$, it is passed through the pre-trained convolution neural network as:\\
\begin{equation}
    \textbf{F}_\mathrm{i} = (\textbf{W}_\mathrm{g} * \mathcal{I}(\mathcal{P}_\mathrm{i}))
\end{equation}
where $\textbf{W}_\mathrm{g}$ represents the overall weights of CNN and * denotes convolution, pooling, and other non-linear functions. The dimension of output feature $\textbf{F}_\mathrm{i}$ is $\mathrm{w}$ x $\mathrm{h}$ x $\mathrm{c}$ where $\mathrm{w}$, $\mathrm{h}$, $\mathrm{c}$ represents the width, height, and channel of the feature map. Note that the CNN in the global stream and the local stream does not share weights.
The feature map $\textbf{F}_\mathrm{i}$ is recurrently passed through different time steps of stacked-LSTMs. The motivation of this step is to make the details finer as the feature map of patch passes through several time steps of LSTMs. So, the input to each time step is the same feature map $\textbf{F}_\mathrm{i}$. The output of the first layer of LSTMs is passed as input to the second layer. The temporal representative function of stacked-LSTMs can be denoted as $\phi$. Hence, the outputs of stacked-LSTMs can be modeled as
\begin{equation}
    [\phi(\textbf{F}_\mathrm{i}^\mathrm{1}),  \phi(\textbf{F}_\mathrm{i}^\mathrm{2}), ...., \phi(\textbf{F}_\mathrm{i}^\mathrm{T}) ]
\end{equation}
where $\mathrm{t}$ = $\mathrm{1}$,$\mathrm{2}$,$\mathrm{3}$ ... $\mathrm{T}$ denotes the time steps of stacked-LSTMs and $\phi
$ denotes the function modelled after each time step by LSTMcell. $\phi(\textbf{F}_\mathrm{i}^\mathrm{t})$ $\in$ $\mathbb{R}^\mathrm{D}$ is the $\mathrm{D}$ dimensional vector denoting  output of feature part( $\mathrm{i}_{th}$ patch) $\textbf{F}_\mathrm{i}$ at time step $\mathrm{t}$. Our experiments \ref{sec:vis} validates our hypothesis about how feature changes over the time steps to focus on finer details of parts.

Once we have finer details of a patch through the LSTM, an attention network is used to perform a weighted aggregation  over these finer features. We believe the advantages of attention is two-fold. First, the trainable weights of attention layer help to provide more weights to the discriminative finer scale of the patch. The attention network helps to focus on the scale of the patch which maximizes the classification accuracy by removing the noisy parts. Secondly, the weighted aggregation of these different time-step features aggregates fine details within the patch. The output of the attention layer can be written as:
\begin{equation}
    \textbf{A}_\mathrm{i} = \sum_{\mathrm{t}=\mathrm{1}}^{\mathrm{T}} \alpha^{\mathrm{t}}\phi(\textbf{F}_\mathrm{i}^\mathrm{t})
\label{eq:atten}
\end{equation}
where
\begin{equation}
    \alpha^{t} = \frac{exp(\textbf{W}^\mathrm{t} \cdot \phi(\textbf{F}_\mathrm{i}^\mathrm{t}))} {\sum_{\mathrm{i}=\mathrm{1}}^{\mathrm{T}} exp(\textbf{W}^\mathrm{t} \cdot \phi(\textbf{F}_\mathrm{i}^\mathrm{t}))}
\label{eq:a2}
\end{equation}where $\textbf{A}_\mathrm{i}$ is the output of attention network and $\textbf{W}^\mathrm{t}$ $\in$ $\mathbb{R}^\mathrm{D}$ is the trainable weight parameter assigned to feature at each time step. 
Finally, the $\mathrm{D}$-dimensional output from attention layer is to pass through a network of fully-connected neural network and softmax to generate class probability vector for  fine-grained categories given by:
\begin{equation}
    \mathcal{\textbf{L}}_\mathrm{I} =  \mathcal{F}^{'}(\textbf{W}_\mathrm{l}  * \textbf{A}_\mathrm{i})
\end{equation}
where $\mathcal{\textbf{L}}_\mathrm{I}$ represents the probability distribution,  $\textbf{W}_\mathrm{l}$ encapsulates the weights of full-connected layer after attention, $\mathcal{F}^{'}$(.) denotes the softmax layer, and  $\textbf{A}_\mathrm{i}$ denotes the output from the attention network. Such design enforces the network to gradually attend to the most discriminative region of patch/part of the image and boost confidence in the prediction of an image.

\subsection{Classification Loss}
The proposed architecture is optimized using classification based loss function. Here, we used two different instances of the same classification loss. So, for a given image the multi-scale loss function can be defined as follows:
\begin{equation}
    \mathcal{L}_\mathrm{total} =   \sum_{\mathrm{n}=\mathrm{1}}^{\mathrm{N}} [\mathcal{L}_{\mathrm{XE}}(Y_\mathrm{n}^\mathrm{g}, Y) + \lambda * \mathcal{L}_{\mathrm{XE}}(Y_\mathrm{n}^\mathrm{l}, Y) ]
\end{equation}
where $\mathcal{L}_{\mathrm{XE}}$ represents classification loss for $N$ training samples. $Y_\mathrm{n}^\mathrm{g}$ denotes predicted label from the probability distribution of global image $\textbf{G}_\mathrm{I}$ and correspondingly $Y_\mathrm{n}^\mathrm{l}$ denotes the  probability distribution of patch representation of local stream $\textbf{L}_\mathrm{I}$ . $Y$ is the ground truth label vector for $n^{th}$ training image. $\lambda$ controls the amount of patch representation's influence on global representation.  The specific classification loss used is the cross-entropy loss given by:
\begin{equation}
    \mathcal{L}_{\mathrm{XE}}(Y_\mathrm{n}^\mathrm{g}, Y) = - \sum_{\mathrm{k}=\mathrm{1}}^{\mathrm{C}} Y^\mathrm{k} log Y_\mathrm{n}^\mathrm{g},
\end{equation}
where $\mathrm{C}$ denotes the total number of classes. Such a design helps the network to learn both global and region-based local patch representative features simultaneously. 

\subsection{Joint Representation}
Once the network is trained end-to-end, we obtain two feature representations of an image $\mathcal{I}$, one from the global stream $\textbf{G}_\mathrm{I}$ and another from the local stream $\textbf{L}_\mathrm{I}$. These descriptors are global and finer part-attention region representations. Hence, to boost the performance we merge the feature output from two-stream to evaluate the performance on the test set. The merge is weighted is the same way as the losses of both streams are weighted.

\section{Experiment Results}
\subsection{Implementation Details}

\paragraph{Datasets} We evaluated the usability and interpretability of our network on the following two datasets:
\begin{itemize}
    \item \textbf{CUB200-2011}\cite{cub} is one of the most used fine-grained classification dataset with 11,788 images from 200 classes. We followed the conventional split with 5,994 training images and 5,794 test images.
    \item \textbf{Stanford Dogs}\cite{dogs} contains 120 breeds of dogs taken from ImageNet. It has 20,580 images from 120 classes with 12000 training images and 8,580 test images.
\end{itemize}

\paragraph{Architectures} We initialize the Convolutional Neural Network of both the stream with ImageNet pre-trained VGG network \cite{1}. We do not use any part annotation or bounding box information. We obtained patches of an image by following the procedure in \cite{rcnn}. Both the streams are trained end-to-end simultaneously. Implementation details of streams are as follows:
\begin{itemize}
    \item \textbf{Global Stream} We have followed the standard practice as per the literature. The input to the global CNN is $448$ x $448$ image. To reduce computation, we removed the fully connected layer from the classifier layer of VGG19 \cite{1} and replace them with Global Average Pooling (GAP) layer \cite{nin}. The classifier layer is a  randomly initialized single fully-connected layer.
    \item \textbf{Local Stream} The output of the weakly supervised network is a set of multiple patches for an image. These patches have varying spatial dimensions. Hence, before passing into local stream's CNN it is resized to $224$ x $224$. Then, the patch is passed through a pre-trained VGG19 \cite{1} network. All the layers after conv5\_4 are removed. Therefore the output of the network is a feature map of $512$ x $14$ x $14$. The feature map is passed through another Global Average Pooling (GAP) layer to output a $512$ -dimensional feature. This feature vector is passed through stacked-LSTMs with a hidden size of $512$. Note that the input feature is the same across all the time-steps of LSTM hence it is computed only once. The number of time steps used is $10$. The output of each step is fed to the attention layer which creates a soft-score based on equation \ref{eq:atten}. These scores are weight-multiplied with LSTM's features and summed to produce a representative feature of the same dimension as hidden layer ($512$). Finally, two fully-connected layers are used to change the $512$ dimensions to the number of classes in datasets ($200$ in CUB200-2011 and $120$ in Stanford dogs). \newline End-to-end training of both streams proceeds with global and local stream having softmax with cross-entropy losses with weight $1.0$ and $1.0$ respectively. At test time, these softmax layers are removed and the prediction is based on the same weighted combination of these two streams. 
\end{itemize}

\subsection{Visualization and Analysis}
\label{sec:vis}
\textbf{Attention Areas:} Insights into the behavior of the local branch can be obtained by visualizing the features of the attention layer and drawing the attention heatmap around the attended regions within the patch. We ran Grad-CAM \cite{gradcam} on the output of the local stream to visualize the finer attended region within the patch. 

\begin{figure}[!ht]
\includegraphics[width=1\linewidth]{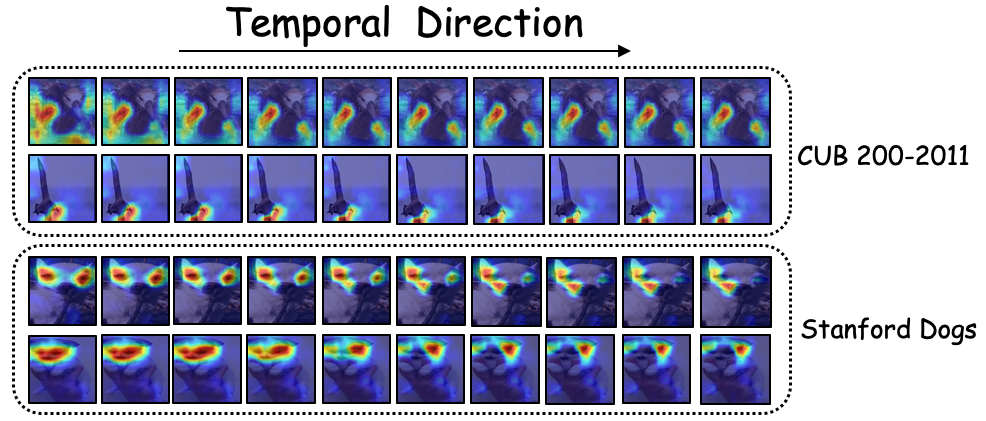}
\caption{The diagram shows the image regions of the patch considered to be important via \cite{gradcam} by each of the 10 hidden representations of LSTMs for its prediction. As evident in the diagram, the attention regions in the patch become finer as we increase the number of time steps from 1 to 10. }
\label{fig:vis}
\end{figure}

\begin{table}[h]
  \centering
    \caption{It shows the accuracy for our network over baseline for CUB200-2011 dataset\cite{cub}}
    \vspace{3mm}
  \label{tab:table1}
    \begin{tabular}{c|c} 
      \hline
      \textbf{Model} &  \textbf{Accuracy(\%)}\\
      \hline \hline
      VGG19 \cite{1} & $\mathrm{77.8}$\\
      \hline
      VGG19 + local-stream  &  $\mathrm{\mathbf{79.6}}$\\
      \hline
    \end{tabular}
\end{table}

The effect of hidden representations of LSTMs from various time-steps is shown in figure \ref{fig:vis}. Using Grad-CAM, \cite{gradcam} we can see the part of the image a time step's hidden representation attends to. Aligning with our motivation we can see that the attention in heatmap goes finer as we go further from the initial time step. As seen in figure \ref {fig:vis} , the hidden representations in initial LSTMs focus on much broader areas of the patch, but as we recurrently pass the patch through the deeper LSTM cells the attention becomes finer and more discriminative. Moreover, in some cases \ref {fig:vis} the attention spans changes from generic regions like the whole face to more subtle variations present in ears, feather, beak. This also shows that the representations at higher time steps are more discriminative producing higher responses. 

Further, the simplicity of the module makes it possible to use it as a plug-n-play module. The local stream can be attached to any network which will be helpful to visualize how the network is attending to the various region of an image. It helps to inject interpretability and get a better understanding of the network evident from figure \ref{fig:vis} . Also, we gain a boost in classification accuracy over the standard baseline as tabulated in Table \ref{tab:table1} for CUB200 dataset and Table in \ref{tab:table2} for Stanford dogs dataset. 

Quantitatively, we tried to analyze the relationship between the level of finer details and being discriminative among the classes in table \ref {tab:table_2}. As evident from the table, the feature representations of finer details become less discriminative as we pass it through more recurrent layers. We processed a single patch repetitively and it started to overfit the finer details.

\begin{table}[h]
  \begin{center}
    \caption{It shows the accuracy for our network over baseline for Stanford Dogs dataset\cite{dogs}}
    \vspace{3mm}
  \label{tab:table2}
    \begin{tabular}{c|c} 
      \hline
      \textbf{Model} &  \textbf{Accuracy(\%)}\\
      \hline \hline
      VGG19 \cite{1} & $\mathrm{77.2}$\\
      \hline
      VGG19 + local-stream  &  $\mathrm{\mathbf{78.7}}$\\
      \hline
    \end{tabular}
  \end{center}
\end{table}

\begin{table}[h]
  \begin{center}
    \caption{Accuracy of fine detail representative feature at different time step $\mathrm{t}$ of LSTM}
    \vspace{3mm}
  \label{tab:table_2}
    \begin{tabular}{c|c} 
      \hline
      \textbf{Feature at time step ($\mathrm{t}$)} &  \textbf{Accuracy(\%)}\\
      \hline \hline
      $\mathrm{1}$ & $\mathrm{78.90}$\\
      \hline
      $\mathrm{2}$ & $\mathrm{79.22}$\\
      \hline
      $\mathrm{3}$ & $\mathrm{\mathbf{79.23}}$\\
      \hline
      $\mathrm{4}$ & $\mathrm{79.11}$\\
      \hline
      $\mathrm{5}$ & $\mathrm{79.04}$\\
      \hline
      $\mathrm{6}$ & $\mathrm{79.09}$\\
      \hline
      $\mathrm{7}$ & $\mathrm{79.08}$\\
      \hline
      $\mathrm{8}$ & $\mathrm{79.03}$\\
      \hline
      $\mathrm{9}$ & $\mathrm{79.03}$\\
      \hline
      $\mathrm{10}$ & $\mathrm{79.04}$\\
      \hline
    \end{tabular}
  \end{center}
\end{table}

\begin{table}[h]
\begin{center}
\caption{Effect of increasing LSTMs time step on classification accuracy for CUB200-2011 dataset\cite{cub}}
\label{tab:ablation}
\resizebox{0.48\textwidth}{!}{
\begin{tabular}{c|c}
\hline
\textbf{Model} &  \textbf{Accuracy(\%)}\\
\hline\hline
VGG19 \cite{1}  & $\mathrm{77.80}$\\
\hline
VGG19 + local-stream(CNN only)  &  $\mathrm{77.79}$ \\
\hline
VGG19 + local-stream(CNN + LSTM) & $\mathrm{78.20}$\\
\hline
VGG19 + local-stream(CNN + LSTM + attention) & $\mathrm{\mathbf{79.60}}$\\
\hline
\end{tabular}}
\end{center}
\end{table}

\subsection{Ablation Study}
We conducted the ablation studies to show how each component individually boost the accuracy of the overall model. 
\begin{table}[h]
  \begin{center}
    \caption{Effect of feature summation vs attention on CUB200-2011 dataset \cite{cub}. '$\sim$' denotes the summation of all the features between specific time steps.}
    \vspace{3mm}
  \label{tab:table10}
    \begin{tabular}{c|c} 
      \hline
      \textbf{Feature Summation} &  \textbf{Accuracy(\%)}\\
      \hline \hline
      $\mathrm{1}$ & $\mathrm{78.90}$\\
      \hline
      $\mathrm{1}$ $\sim$ $\mathrm{2}$ & $\mathrm{78.78}$\\
      \hline
       $\mathrm{1}$ $\sim$ $\mathrm{3}$ & $\mathrm{79.18}$\\
      \hline
      $\mathrm{1}$ $\sim$ $\mathrm{4}$ & $\mathrm{78.75}$\\
      \hline
      $\mathrm{1}$ $\sim$ $\mathrm{5}$ & $\mathrm{77.04}$\\
      \hline
      $\mathrm{1}$ $\sim$ $\mathrm{6}$ & $\mathrm{75.32}$\\
      \hline
      $\mathrm{1}$ $\sim$ $\mathrm{7}$ & $\mathrm{72.97}$\\
      \hline
      $\mathrm{1}$ $\sim$ $\mathrm{8}$ & $\mathrm{71.01}$\\
      \hline
      $\mathrm{1}$ $\sim$ $\mathrm{9}$ & $\mathrm{69.21}$\\
      \hline
      $\mathrm{1}$ $\sim$ $\mathrm{10}$ & $\mathrm{67.64}$\\
      \hline
      $\mathrm{Attention}$ & $\mathrm{\mathbf{79.60}}$\\
      \hline
    \end{tabular}
  \end{center}
\end{table}

\textbf{Effect of network components on classification} As shown in Table \ref{tab:ablation}, the presence of only Convolutional Neural Network in the local-stream doesn't add much performance benefit. Further, a stacked-LSTM layer is added in the local-stream. Here, the local-stream is trained using cross-entropy losses on the outputs of all the time steps. During inference, we only consider the output of the final step. This addition of the stacked-LSTM layer boosts the performance by a significant margin ($\sim$1\%) , indicating the finer details are highly discriminative. Moreover, the attention layer provides extra gain to reach much better performance showing the effectiveness of weighted aggregation of the finer features.

\textbf{Attention vs Summation} We investigate the effect and importance of attention in the local-stream of the network, we tried to replace the attention layer with a simple summation of features. Table \ref{tab:table10} shows the result of an experiment comparing simple summation of features from time step $\mathrm{1}$ to $\mathrm{10}$ with the attention layer. The results validate the claim that simple summing doesn't help to boost the accuracy while the attention layer explicitly learns the weights for each feature at the time steps. This helps the network to focus on the finer details which is most discriminative. 

\subsection{Hyperparameter Setting For Time Steps}
We tried to see how the number of steps affects the overall classification accuracy of the network. We ran the network on CUB200-2011 dataset with different number of time steps in each run and recorded the results in Table \ref{tab:table3}
\begin{table}[h]
\begin{center}
\caption{Effect of increasing LSTMs time step on the classification accuracy for CUB200-2011 dataset\cite{cub}}
\label{tab:table3}
\vspace{3mm}
\resizebox{0.25\textwidth}{!}{
\begin{tabular}{c|c} 
\hline
\textbf{Time Steps} &  \textbf{Accuracy(\%)}\\
\hline\hline
$\mathrm{5}$  & $\mathrm{77.72}$\\
\hline
$\mathrm{10}$  &  $\mathrm{\mathbf{79.60}}$\\
\hline
$\mathrm{15}$ & $\mathrm{79.12}$\\
\hline
$\mathrm{20}$ &  $\mathrm{79.34}$\\
\hline
$\mathrm{25}$ & $\mathrm{78.98}$\\
\hline
\end{tabular}}

\end{center}
\end{table}
This shows that adding more time step does not necessarily increase the performance of the network. The results also align well with figure \ref{fig:vis} showing diminishing difference in fine attention towards the end of the recurrent time steps.

\section{Conclusion}
In this paper, we propose a simple recurrent attention based module that extracts finer details from the image providing more discriminative features for fine-grained classification. The local stream of whole architecture aggregates these fine details into a representative and complementary feature vector. The proposed method does not need bounding box/part annotation for training and can be trained end-to-end. Moreover, the simplicity of the module makes it a plug-n-play module, thus, increasing its usability. Through the ablation study, we also show the effectiveness of each part of the network. Additionally, the interpretable nature of the module makes it easy to visualize learned discriminative patches.

{\small
\bibliographystyle{ieee_fullname}
\bibliography{egbib}
}

\end{document}